\documentclass{article} % For LaTeX2e
\usepackage{lmrl2025_workshop,times}
\lmrlfinalcopy

\usepackage{amsmath,amsfonts,bm}

\def\eqref#1{equation~\ref{#1}}
\def\1{\bm{1}}

\DeclareMathAlphabet{\mathsfit}{\encodingdefault}{\sfdefault}{m}{sl}
\SetMathAlphabet{\mathsfit}{bold}{\encodingdefault}{\sfdefault}{bx}{n}

\usepackage{hyperref}
\usepackage{url}
\usepackage{booktabs}

\title{To Bin or not to Bin: Alternative Representations of Mass Spectra}

\author{Niek de Jonge, Justin J. J. van der Hooft\thanks{Department of Biochemistry, University of Johannesburg, South Africa}, Daniel Probst\\
Bioinformatics Group\\
University \& Research Wageningen\\
The Netherlands\\
\texttt{\{niek.dejonge,justin.vanderhooft,daniel.probst\}@wur.nl} \\
}

\begin{document}

\maketitle

% \begin{abstract}
% Mass spectra are widely used in the life sciences and beyond, often representing molecules with unknown molecular structure. Hence, predicting molecular properties or the relation between molecules directly from mass spectra is of great interest. However, to create uniform vector sizes and remove noise, they are generally either binned, tokenized, or sub-sampled for comparison or featurization. Here, we investigate set and graph representations as alternatives that avoid any of these preprocessing steps, retaining all m/z and intensity values across mass spectra. We present preliminary results that show the validity of representing mass spectra as sets and graphs, and provide examples of set- and graph-based baselines outperforming an established binning-based baseline.
% \end{abstract}

\section{Introduction}
Mass spectrometry, especially so-called tandem mass spectrometry, is commonly used to assess the chemical diversity of samples. The resulting mass fragmentation spectra are representations of molecules of which the structure may have not been determined. This poses the challenge of experimentally determining or computationally predicting molecular structures from mass spectra. An alternative option is to predict molecular properties or molecular similarity directly from spectra. Various methodologies have been proposed to embed mass spectra for further use in machine learning tasks. However, these methodologies require preprocessing of the spectra, which often includes binning or sub-sampling peaks with the main reasoning of creating uniform vector sizes and removing noise.

Here, we investigate two alternatives to the binning of mass spectra before down-stream machine learning tasks, namely, set-based and graph-based representations. Comparing the two proposed representations to train a set transformer and a graph neural network on a regression task, respectively, we show that they both perform substantially better than a multilayer perceptron trained on binned data.

\section{Related Work}
Machine learning models commonly take binned mass spectra as input, discretise peaks as part of a vocabulary through tokenisation, or select peaks with top-$n$ intentsities. Examples are MS2DeepScore by~\citet{huberMS2DeepScoreNovelDeep2021}, Spec2Vec by~\citet{huberSpec2VecImprovedMass2021}, MS2Mol by~\citet{butlerMS2MolTransformerModel2023}, MSBert by~\citet{zhangMSBERTEmbeddingTandem2024}, and DreaMS by~\citet{bushuievEmergenceMolecularStructures2025}. Meanwhile, models targeting the reverse problem, predicting mass spectra from molecular graphs, often apply graph neural networks or graph transformers to the input molecular graphs. Recent examples include MassFormer by~\citet{youngTandemMassSpectrum2024} and others~\citep{zhangPredictionElectronIonization2022,murphyEfficientlyPredictingHigh2023,parkMassSpectraPrediction2024}. 

Recently, ~\citet{nallapareddyImprovingFulllengthRibosome2024} have represented sequential mRNA codons as graphs, inspiring us to view a mass spectrum as a sequence of intensities along the mass-to-charge ratio dimension. Furthermore, applying graph neural networks to time series has become a commonly used approach, hinting at the potential of GNNs applied to mass spectra~\citep{jinSurveyGraphNeural2024}.

The representation of mass spectra as sets is based on a recent publication on set representation learning for molecules by~\citet{boulougouriMolecularSetRepresentation2024}. As peaks in mass spectra are pairs of intensities and m/z values of unequal numbers across spectra, a mass spectrum can be viewed as a set of intensity-m/z value pairs.

\section{Methodology}

Data and splits provided by~\citet{jongeBridgingPolaritiesMetabolomics2025} were used for the experiments. The quantitative estimate of drug-likeness (QED) values were calculated using the RDKit implementation of the metric~\citep{landrumrdkit}. While the data set provides additional metadata, only the m/z (including the precursor) and intensity values were used for all subsequent steps. Further processing depended on the downstream architecture. For the multilayer perceptron (MLP), the m/z peaks were binned following the protocol provided by~\citet{huberMS2DeepScoreNovelDeep2021}, changing the range of bins from 10 to 10,000 to 0 to 10,000) to have consistency across experiments. For the SetTransformer, m/z--intensity pairs were encoded as a PyTorch TensorDataset. Finally, the data was encoded as set of graphs for the GNN: Each peak in a mass spectrum is represented by a vertex and connected to neighboring peaks through edges; the intensities were used as vertex attributes and the differences in m/z as edge attributes. An additional vertex with m/z and intensity of 0 was created to encode the m/z delta of the initial peak. For all experiments, intensities were normalized and the precursor m/z was treated as a normal peak with an intensity of 2.0.

To evaluate the binned data, an MLP with 2 hidden layers (1,024 and 512 neurons, respectively) with ReLU activation and a dropout of 0.5 was used. The SetTransformer, trained on the sets of m/z--intensity pairs, was configured with two hidden layers (32 and 16 neurons, respectively). Finally, the graph neural network, a graph attention network (GAT), was trained with 8 message passing layers, 1,024 hidden channels, and global mean pooling. The hyperparameters of the GAT were chosen to result in a similar number of parameters to the MLP and not further optimized.

\section{Results and Discussion}
We compare the three different methods using binned, set, and graph representations of mass spectra as input on the regression task of predicting the QED of a molecule from its mass spectra~\citep{bickertonQuantifyingChemicalBeauty2012}. As QED is a broad composite descriptor capturing both molecular structure and biological function, it serves as an effective surrogate for assessing the proficiency of a methodology in predicting molecular properties from mass spectrometry data through regression analysis.

\begin{table}[h]
\centering
\caption{Performances of models trained on different mass spectrum representations. The MLP was trained on binned mass spectra, the SetTransformer on sets of m/z-intensity pairs, and the GNN (GAT) on graphs where the vertices represent peaks (intensity), and the edges distances (m/z) between peaks. Over all metrics, the GNN-based approach performs best.}
\vspace{5pt}
\label{tbl:perf}
\begin{tabular}{lrrrrr}
\toprule
Model & Params & MAE ($\downarrow$) & RMSE ($\downarrow$) & Pearson's \textit{r} ($\uparrow$) & R\textsuperscript{2} ($\uparrow$) \\
\midrule
MLP         & 11.0 M & 0.145 $\pm$ 0.008 & 0.200 $\pm$ 0.008 & 0.736 $\pm$ 0.011 & 0.437 $\pm$ 0.043 \\
SetTransformer & 0.4 M & 0.134 $\pm$ 0.002 & 0.174 $\pm$ 0.001 & 0.758 $\pm$ 0.004 & 0.572 $\pm$ 0.006 \\
GNN (GAT)          & 12.6 M & \textbf{0.110 $\pm$ 0.006} & \textbf{0.144 $\pm$ 0.006} & \textbf{0.843 $\pm$ 0.015} & \textbf{0.709 $\pm$ 0.025} \\
\bottomrule
\end{tabular}
\end{table}

Our experiments, with results shown in Table \ref{tbl:perf}, yielded three primary results: (1) Both the set representation and graph representation with their related architectures performed better than the binned representation-trained MLP. (2) The set representation model is highly efficient with only 400 k parameters, compared to 11.0 M and 12.6 M parameters of the other two models. (3) Representing mass spectra as graphs results in better performance than both binning and set representation. Given the performance increase of the GNN compared to the set-based approach, both of which use all available data without binning or sub-sampling, hints at the possibility that graphs are a favorable representation of mass spectra, as meaningful information may be propagated along the graph. As we chose a relative simple regression task, further research is needed to assess the proposed architectures on other tasks that are common in metabolomics research, including similarity prediction or molecular structure prediction. 

\section{Conclusion}

We showed that representing mass spectra as sets or graphs is not only possible but performs better in our prospective study compared to the commonly used approach of representing mass spectra as fixed-length arrays by binning or sub-sampling peaks. As both established and recent machine learning approaches rely on binning or sub-sampling of mass spectra~\citep{huberMS2DeepScoreNovelDeep2021,bushuievEmergenceMolecularStructures2025}, we believe that our results are of immediate interest for machine learning researchers developing new architectures applied to mass spectra data. Furthermore, we provide ready-to-use encoders and models that can easily be integrated in existing machine learning architectures used in mass spectra-related tasks. Finally, the code and the scripts to get the data used in this study can be found in the following repository: \url{https://anonymous.4open.science/r/massgraph-C84F}.

\bibliography{preprint}
\bibliographystyle{iclr2025_conference}

\end{document}